\pdfoutput=1
\documentclass[10pt,twocolumn,letterpaper]{article}

\usepackage[pagenumbers]{iccv}

\usepackage{tcolorbox}
\newtcolorbox{keybox}{
  colback=black!4, colframe=black!22, boxrule=0.5pt, arc=2pt,
  left=6pt, right=6pt, top=4pt, bottom=4pt,
  before skip=\topsep, after skip=\topsep,
}

\definecolor{iccvblue}{rgb}{0.21,0.49,0.74}
\usepackage[pagebackref,breaklinks,colorlinks,allcolors=iccvblue]{hyperref}

\title{Medical Image Segmentation Is Now Just Another Coding Task}

\author{
Shuheng Cao$^{1,*}$\quad
Ruiqi Chen$^{2,*}$\quad
Qizhen Lan$^{3}$\quad
Edward Wang$^{1}$\quad
Xiyun Hu$^{4}$\\
Yilei Wang$^{1}$\quad
Juqy Chen$^{1}$\quad
Weilin Huang$^{1}$\quad
Anhong Guo$^{2,\dagger}$\\[0.6em]
{\normalsize $^{1}$University of California, San Diego\quad $^{2}$University of Michigan}\\
{\normalsize $^{3}$The University of Texas Health Science Center at Houston\quad $^{4}$University of North Carolina at Chapel Hill}\\[0.3em]
{\tt\small ruiqich@umich.edu}
}

\begin{document}
\maketitle
{\renewcommand{\thefootnote}{\fnsymbol{footnote}}\footnotetext[1]{Equal contribution.\quad $^{\dagger}$Corresponding author.}}
\begin{abstract}
For decades, every new medical image segmentation task has needed an expert, a dedicated architecture, and a framework such as nnU-Net. This is no longer the case: medical image segmentation is now just another coding task. Given only a task description, labeled data, and a CPU---with no segmentation framework, no pretrained weights, and no human in the loop---four off-the-shelf coding agents built segmenters for 13 tasks across four imaging modalities within hours, an ability that has emerged rather than been built. On the nine 2D tasks, their best programs reach 99\% of the Dice of a GPU-trained nnU-Net. Like any code, a segmenter improves by being revised and checked, so we ask how far the same agents can improve their own segmentations and what unlocks further gains. On unseen cases there is no reference, so the agents inspect renderings of their own predictions: this alone lifts the worst tenth of cases by 0.09 Dice and cuts boundary error (HD95) by almost a quarter. Far more is within reach. Told the true Dice of each attempt, the same agents improve in all 14 matched pairs, rescue 85\% of failed masks instead of 30\%, and surpass the GPU-trained nnU-Net on average, by producing better masks rather than by choosing among them. What unlocks these gains is a quality signal, which neither visual inspection, nor more rounds, nor a quality estimator trained by the agents themselves yet supplies: we call this the \emph{quality blind spot}. The ability to build and to repair is already present; the next advance will come not from another segmentation model, but from reference-free quality estimation that coding agents can call.
\end{abstract}
\section{Introduction}
\label{sec:intro}

\begin{figure*}[t]
\centering
\includegraphics[width=\textwidth]{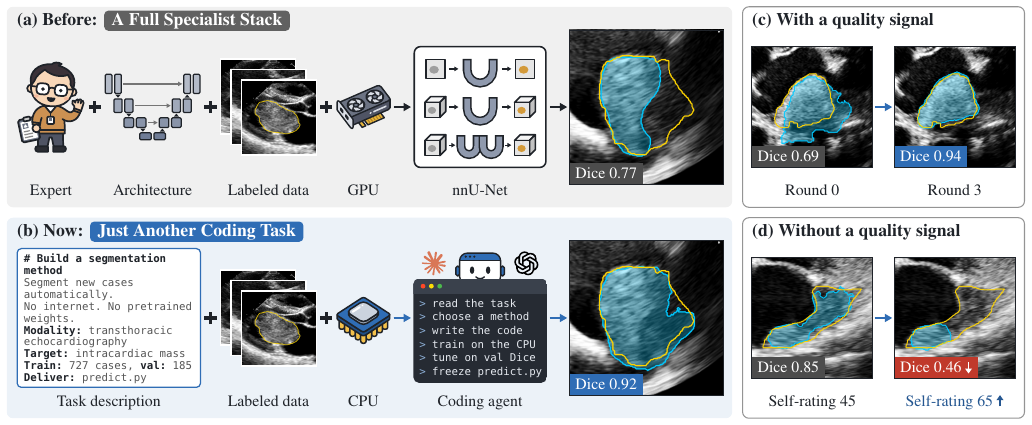}
\caption{\textbf{From a specialist stack to a coding task.}
\textbf{(a)~Before:} a modality expert, dedicated architecture, labeled data, and a GPU feed a self-configuring framework, nnU-Net~\cite{isensee_nnu-net_2020}.
\textbf{(b)~Now (Build):} an unmodified coding agent (Claude Code or Codex CLI) takes only a task description, labeled data, and a CPU to deliver its own segmenter, lifting the exact same test case (CardioMassTTE-1151~\cite{zhang2026cardiomasstte}; yellow: reference, cyan: prediction) from 0.77 to \textbf{0.92} Dice and reaching \textbf{99\%} of nnU-Net Dice across 2D tasks.
\textbf{(c)~With a quality signal:} told scalar True-Dice after each round, the same agent repairs a flawed mask from 0.69 to \textbf{0.94} Dice in three rounds.
\textbf{(d)~Without a quality signal (quality blind spot):} visual self-review raises its 0--100 self-rating from 45 to 65 even as Dice drops from 0.85 to 0.46---\textbf{seeing is not verifying}.}
\label{fig:overview}
\end{figure*}

For decades, segmenting a new class of medical images has meant assembling a full \emph{specialist stack}: a modality expert; a dedicated architecture~\cite{ronneberger2015unetconvolutionalnetworksbiomedical}; a self-configuring framework such as nnU-Net~\cite{isensee_nnu-net_2020} encoding years of heuristics for preprocessing, training, and post-processing; GPUs for training; and, more recently, foundation models pretrained on large mask collections~\cite{MedSAM}. Every piece of this stack was purpose-built for segmentation, and each new task or modality requires standing it up anew.

This paper shows that a general-purpose coding agent needs none of it. Given only a task description, labeled data, and a CPU---with no segmentation framework, no pretrained weights, no domain skills, and no human in the loop---four off-the-shelf coding agents autonomously choose a method, write preprocessing and training code, tune against validation Dice, and deliver a finished segmentation program within hours across 13 tasks and four imaging modalities (\Cref{fig:overview}). On the nine 2D tasks, their best programs reach \textbf{99\%} of the Dice (a 0--1 mask-overlap score) of a GPU-trained nnU-Net (surpassing it outright on three tasks) with a lower failure rate (\textbf{24.0\%} vs.\ 25.8\%). Model training has not disappeared: it has become one step inside a program that the agent writes and runs itself, like any other code. That is what our title means: \emph{medical image segmentation is now just another coding task}---an \emph{emergent} capability for which these agents were never designed or trained~\cite{wei2022emergentabilitieslargelanguage, schaeffer2023emergentabilitieslargelanguage}.

Once segmentation is a coding task, improving a segmenter becomes a coding agent problem as well: \emph{revise and check}~\cite{jimenez2024swebenchlanguagemodelsresolve, yang2024sweagentagentcomputerinterfacesenable}. During development, that check is validation Dice; on unseen test cases, there is no reference. We therefore ask two questions: \textbf{How much can the same agents improve their own predictions here, and what unlocks further gains?}
By inspecting visual renderings of their own predictions case by case (Self-review), the agents lift the worst 10\% of cases by \textbf{0.09 Dice} and cut boundary error (HD95) by \textbf{24\%}. Far more headroom is within reach. Across 14 matched program--task pairs starting from the exact same frozen programs and test cases, telling the agent only the true scalar Dice of its mask after each round (True-Dice: a 0--1 overlap score without revealing the reference mask, analogous to a unit-test score) improves every pair in every round, lifts the failed-mask rescue rate from 30\% to \textbf{85\%}, cuts the failure rate to one-third of nnU-Net's (\textbf{4.1\%} vs.\ 11.9\%), and surpasses GPU-trained nnU-Net on average (\textbf{0.790} vs.\ 0.765 Dice).

What unlocks this headroom is a \emph{quality signal}: a score telling the agent whether a revision made the mask better or worse. Because no reference is available at deployment, we test the three most natural substitutes: the agent's own visual judgment (which believes \textbf{91\%} of modified cases improved when only \textbf{50\%} actually did), additional rounds of revision ($+$0.004 Dice over Rounds~2--5 without a signal vs.\ \textbf{{\boldmath$+$}0.119} with True-Dice), and a reference-free quality estimator trained by the agent itself (\textbf{0\%} median gap closed). We call this phenomenon the \emph{quality blind spot}: without a reference, the agent can see which cases are hard, but cannot tell whether its own revision made the mask better or worse. That agents need feedback to improve is known in software~\cite{jimenez2024swebenchlanguagemodelsresolve, yang2024sweagentagentcomputerinterfacesenable, huang2024largelanguagemodelsselfcorrect, kamoi-etal-2024-llms}; our new finding is that vision cannot supply it: \textbf{seeing is not verifying}. Segmentation inherits both halves of a coding task: the agent can build the program, and to improve it, the agent needs a signal. The next advance will therefore come not from another segmentation model, but from reference-free quality estimation that coding agents can call.

Our contributions are threefold:
\begin{itemize}
    \item \textbf{Segmentation as a coding task.} Unmodified coding agents given only a task description, labeled data, and a CPU build segmenters across 13 tasks and four controlled tool settings, reaching \textbf{99\%} of GPU-trained nnU-Net Dice on 2D tasks without any specialist stack.
    \item \textbf{Self-improvement with large headroom.} Visual self-review lifts the worst 10\% of unseen cases by 0.09 Dice, while supplying true Dice at each round improves all 14 matched pairs, rescues \textbf{85\%} of failed masks, and surpasses nnU-Net on average by generating better masks across overlap, boundary, size, and reliability.
    \item \textbf{The quality blind spot.} Neither visual inspection (\emph{seeing is not verifying}), nor extra rounds, nor self-trained estimators supplies this needed quality signal, pointing the next advance toward reference-free quality estimation.
\end{itemize}
\section{Related work}
\label{sec:related_work}

\noindent\textbf{Agents that build segmenters.}
Prior efforts to automate medical segmentation with agents all add a layer on top of the specialist stack: Camyla~\cite{gao2026camylascalingautonomousresearch} wraps an agent around a GPU-trained nnU-Net framework; Moris and Orlando~\cite{moris2026codingagentsbuildrobust} craft task-specific Claude Code skills with human stage-by-stage researcher review; Simple Agents~\cite{xuefei2025simpleagentsoutperformexperts} optimizes pre- and post-processing around a fixed segmentation tool; Arjomandi et al.~\cite{arjomandi2026prompts} and Zhao et al.~\cite{zhao2026fullyautomatedmedicalimaging} prompt or steer models for medical imaging pipelines executed or validated via code; and systems such as AMID~\cite{liu2026autonomousauditablemedicalimaging} and ReX-MLE~\cite{kenia2025rexmleautonomousagentbenchmark} evaluate domain-specific multi-agent frameworks on medical imaging challenges (where autonomous agents largely trail human experts~\cite{kenia2025rexmleautonomousagentbenchmark}). Our study works by subtraction: unmodified, general-purpose coding agents run on a CPU with no segmentation framework, no pretrained weights, and no human intervention (including a classical-only setting reminiscent of evolutionary operator pipelines~\cite{cortacero2023evolutionary}), isolating the coding agent itself while using nnU-Net~\cite{isensee_nnu-net_2020} and MedSAM~\cite{MedSAM} only as a reference benchmark and an optional tool setting.

\smallskip
\noindent\textbf{Agents that inspect and revise masks.}
Taurus~\cite{hu2026taurusllmintheloopctannotation}, nnInteractive Agent~\cite{isensee2026nninteractiveagentgeneralpurposeai}, and MedSAM-3 Agent~\cite{liu2026medsam3delvingsegmentmedical} let an agent inspect renderings and revise masks produced by external tools. Beyond reporting overall gains~\cite{hu2026taurusllmintheloopctannotation}, we quantify across 13 tasks exactly what visual review fixes---the hardest cases, boundary error, and size error via higher precision. Crucially, our agent reviews unseen predictions of the frozen program that \emph{it built itself from scratch}, isolating self-repair from external tool assistance and bounding it against true-Dice feedback.

\smallskip
\noindent\textbf{Self-correction and verification.}
Coding agents on software rely on executable feedback~\cite{jimenez2024swebenchlanguagemodelsresolve, yang2024sweagentagentcomputerinterfacesenable}, LLM self-correction on text relies largely on oracles~\cite{huang2024largelanguagemodelsselfcorrect, kamoi-etal-2024-llms}, visual grounding studies compare intrinsic, oracle, and verifier feedback~\cite{liao2025largevisionlanguagemodelscorrect}, and test-time compute scaling shows that extra iterations help only when guided by a reliable verifier~\cite{snell2024scalingllmtesttimecompute} rather than an imperfect proxy~\cite{gao2022scalinglawsrewardmodel}; meanwhile, existing reference-free segmentation quality estimators~\cite{mossina2025conformalpredictionimagesegmentation, senbi2024groundtruthfreeevaluationsegmentationmedical} focus on global scoring rather than ranking competing revisions of the same case. We conduct the \textbf{first} matched true-Dice comparison on segmentation masks, decomposing gains into generation versus selection and testing whether direct visual inspection can replace a reference. Whereas an oracle in text tasks mainly helps a model stop at the right attempt~\cite{huang2024largelanguagemodelsselfcorrect}, on segmentation it enables the agent to generate masks it would otherwise never produce---yet without it, even direct visual inspection cannot substitute for verification (\emph{seeing is not verifying}).
\section{Study design}
\label{sec:study_design}
Our study proceeds in three stages, strictly controlling what the coding agent receives at each step (\Cref{fig:overview}). In the \emph{build} stage, the agent writes a segmentation program from labeled data, which is then frozen. In the \emph{self-review} stage, the same agent visually reviews the test-set predictions of its frozen program case by case without a reference. In the \emph{matched re-review} stage, we start from the exact same frozen program and test cases, varying one condition at a time to measure how much further the agent can improve and what unlocks those gains.

\subsection{Tasks and metrics}
\label{sec:design_tasks}
We evaluate across 13 medical image segmentation tasks (nine 2D and four 3D) drawn from 9 public datasets spanning X-ray, MRI, ultrasound, and CT (\Cref{tab:build_results}). 8 of the 13 tasks come from 5 datasets released publicly only after the training cutoff of every evaluated agent. All datasets are strictly anonymized with neutral identifiers and stripped metadata headers.

The unit of evaluation is a single mask, defined as one anatomical region of one test case. We evaluate each mask across a four-dimensional metric panel: \textbf{Overlap} via the Dice similarity coefficient; \textbf{Boundary} via the 95th-percentile Hausdorff distance (HD95); \textbf{Size} via relative volume difference ($\text{RVD} = |V_{\text{pred}} - V_{\text{ref}}| / V_{\text{ref}}$, using area in 2D); and \textbf{Reliability} via \emph{failure rate} (the fraction of masks with $\text{Dice} < 0.5$) and \emph{worst-10\% Dice} (mean Dice of the decile of masks with the lowest initial Dice). A failed mask reaching $\text{Dice} \ge 0.5$ after review is \emph{rescued}.

\subsection{Agents, tool settings, and the reference model}
\label{sec:design_agents}

We evaluate four unmodified, off-the-shelf coding agents across two vendors: Claude Opus~5.5 and Claude Fable~5.1 (in Claude Code), and GPT-6 Astra and GPT-6.1 Sol (in Codex CLI). Every agent receives the exact same task-agnostic prompt with no domain-specific skills, no external tool servers, and no human intervention. All agent-written code executes on a CPU inside a network-isolated sandbox.

We enforce four controlled tool settings via the execution environment across 91 runs, where a \emph{run} is one complete execution of one agent on one task under one setting. Setting~2 (39 runs, 8\,h; neural networks trained from scratch without segmentation frameworks or pretrained weights) is our primary condition; Setting~1 (40 runs, 4\,h; classical methods only) tests whether building a segmenter requires deep learning at all; and Settings~3 and~4 (6 runs each, 4\,h and 8\,h) test pretrained MedSAM and MedSAM2 with agent-generated box prompts, either frozen or with fine-tuning permitted. The specialist reference model is nnU-Net v2.8.1~\cite{isensee_nnu-net_2020}, trained from scratch on each task on a GPU.

\subsection{Build: development and freezing}
\label{sec:design_build}

Given a brief task description alongside labeled training and validation sets, the agent freely explores any algorithmic approach and checks iterations against validation Dice. At the end of the session, it delivers a deterministic inference program, which is frozen and run across all test cases to produce initial predictions.

\subsection{Self-review: reference-free, case by case}
\label{sec:design_self_review}

Next, the same agent reviews the predictions of its own frozen program case by case on a fixed subset of 50 test cases per task. Each case is handled by an isolated sub-agent session that inspects contour-overlay renderings of the current prediction alongside the raw image (side-by-side in 2D; multi-planar slice montages in 3D). Strictly without access to the reference mask or any true score, the sub-agent may re-run the frozen program with modified parameters or write code to edit the mask directly for up to 5 rounds, recording a 0--100 visual self-rating each round before selecting its final submission.

\subsection{Matched re-review: one upper bound, three candidates}
\label{sec:design_matched}

To measure how much headroom remains and what unlocks it, we start from the identical frozen program and the same 20 test cases per pair, launching fresh 5-round review sessions with the exact same editing tools while varying a single condition at a time across 14 matched program--task pairs (spanning 10 tasks):
(1)~\textbf{Self-review} (14 pairs) gives the agent only its own eyes, measuring gains from visual inspection alone;
(2)~\textbf{True-Dice} (14 pairs) returns the true scalar Dice per label after each round without revealing the reference mask, serving solely as a diagnostic upper bound to quantify how much a perfect scalar quality signal can unlock;
(3)~\textbf{Forced-rounds} (4 tasks) withholds all external scores but requires the agent to complete all 5 rounds on every case, testing whether extra revision effort can substitute for a signal; and
(4)~\textbf{Self-critic} (7 runs) lets the agent first spend 60 minutes training its own reference-free Dice estimator on labeled training/validation predictions and synthetic mask perturbations, then query it to guide each test-case review round.
\section{Coding agents build specialist-level segmenters}
\label{sec:build}

\begin{table*}[t]
\caption{\textbf{Build-stage performance: frozen agent programs vs.\ GPU-trained nnU-Net.} Left: mean test-set Dice per task under Setting~1 (classical only) and Setting~2 (neural networks trained from scratch on a CPU), with macro-averaged Dice and percentage of nnU-Net across tasks. Right: mask-by-mask comparison of Setting~2 best programs against nnU-Net on all test sets (failure: Dice $<0.5$). $^*$Dataset released publicly after the training cutoff of all evaluated agents.}  \label{tab:build_results}
  \centering
  \footnotesize
  \setlength{\tabcolsep}{5pt}
  \begin{tabular}{@{}lccccccc@{}}
    \toprule
    & \multicolumn{4}{c}{Mean test Dice}
    & \multicolumn{3}{c}{Setting~2 best, mask by mask} \\
    \cmidrule(lr){2-5} \cmidrule(l){6-8}
    & & Setting~1 & \multicolumn{2}{c}{Setting~2}
    & Masks within 0.05 & \multicolumn{2}{c}{Failure rate} \\
    \cmidrule(lr){4-5} \cmidrule(l){7-8}
    Task & nnU-Net & best & median & best
    & Dice or better & Agent & nnU-Net \\
    \midrule
    \multicolumn{8}{@{}l}{\textbf{2D tasks}} \\
    CARDIAG vessels$^*$~\cite{lau2026cardiagdensesegmentclassification} & .876 & .812 & .843 & .850          & 86\% & \textbf{\phantom{0}0.7\%} & \phantom{0}0.7\% \\
    CARDIAG segments$^*$ & .381 & .255 & .311 & .328          & 58\% & 47.0\%          & 39.3\% \\
    CARDIAG stenosis$^*$ & .090 & .052 & .123 & \textbf{.152} & 90\% & \textbf{74.4\%} & 93.9\% \\
    ARCADE segments~\cite{popov2024dataset} & .356 & .190 & .328 & .354          & 69\% & 36.0\%          & 34.7\% \\
    ARCADE stenosis & .560 & .370 & .458 & \textbf{.576} & 65\% & \textbf{28.3\%} & 33.0\% \\
    FeEcho4D$^*$~\cite{hasan_2026_21322299} & .719 & .749 & .756 & \textbf{.757} & 97\% & \textbf{\phantom{0}7.4\%} & 10.2\% \\
    CardioMassTTE-1151$^*$~\cite{zhang2026cardiomasstte} & .804 & .483 & .747 & .766          & 72\% & \phantom{0}8.1\%          & \phantom{0}6.4\% \\
    CAMUS~\cite{leclerc2019camus} & .921 & .876 & .913 & .916          & 97\% & \textbf{\phantom{0}0.0\%} & \phantom{0}0.0\% \\
    CesareanScarDefect$^*$~\cite{tian2026cesareanscardefect} & .740 & .658 & .685 & .718          & 74\% & \textbf{13.9\%} & 13.9\% \\
    \midrule
    \textbf{2D mean}       & .605 & .494 & .574 & \textbf{.602} & \textbf{79\%} & \textbf{24.0\%} & 25.8\% \\
    \textbf{\% of nnU-Net} & ---  & 82\% & 95\% & \textbf{99\%} &               &                 &        \\
    \midrule
    \multicolumn{8}{@{}l}{\textbf{3D tasks}} \\
    WAW-GliomAI pre-op$^*$~\cite{bartnik2026wawgliomai} & .860 & .744 & .646 & .684 & 42\% & 18.5\% & \phantom{0}7.3\% \\
    WAW-GliomAI post-op$^*$ & .623 & .546 & .468 & .502 & 36\% & 29.4\% & 21.5\% \\
    MSD BrainTumour~\cite{simpson2019msd,antonelli2022msd} & .786 & .730 & .729 & .733 & 57\% & 13.2\% & 10.0\% \\
    CT-PEGCT-Diag~\cite{zhou2026ctpegctdiag} & .846 & .722 & .703 & .738 & 50\% & 17.2\% & \phantom{0}9.4\% \\
    \midrule
    \textbf{3D mean}       & .779 & .685          & .637 & .664 & 46\% & 19.6\% & 12.0\% \\
    \textbf{\% of nnU-Net} & ---  & \textbf{88\%} & 82\% & 85\% & & & \\
    \bottomrule
  \end{tabular}
\end{table*}

This section answers the first question: \textit{why medical image segmentation is now just another coding task}. We evaluate the frozen programs delivered directly at the end of the build stage, prior to any case-by-case review.

\subsection{Specialist-level accuracy on 2D tasks}
\label{sec:build_2d}

\Cref{tab:build_results} compares the programs written autonomously by the coding agents on a CPU against a GPU-trained nnU-Net. Across the nine 2D tasks, the best Setting~2 programs achieve a mean Dice of 0.602 vs.\ 0.605 for nnU-Net---reaching \textbf{99\%} of the specialist framework's performance without a single component of the specialist stack. Eight tasks are within 0.04 Dice of nnU-Net, five are within 0.01 Dice or better, and three (CARDIAG stenosis, ARCADE stenosis, and FeEcho4D) surpass nnU-Net outright. Nor does this parity depend on cherry-picking the best run: the median Setting~2 run across agents achieves a mean Dice of 0.574, or \textbf{95\%} of nnU-Net. Every program was written from scratch through the agent's standard coding loop---turning what once required a domain expert and a dedicated segmentation stack into the routine workflow of a general-purpose coding agent. On the four volumetric 3D tasks, the same CPU-bound workflow already reaches 85\%--88\% of GPU-trained nnU-Net Dice within a fixed 4--8\,h compute envelope (\Cref{tab:build_results}, bottom; \Cref{sec:build_cost}).

\begin{figure*}[t]
  \centering
  \includegraphics[width=\textwidth]{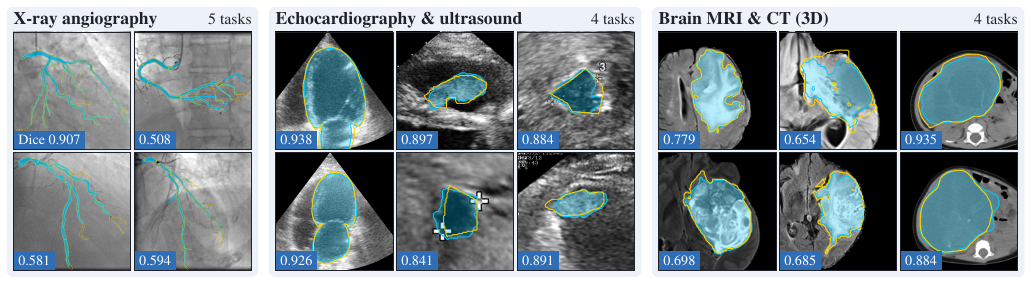}
  \caption{\textbf{Zero-touch generalization of CPU-built agent programs across modalities and dimensions.} Autonomous Setting~2 programs segment unseen test cases directly out of the box (yellow: reference, cyan: prediction; tag: Dice, computed over the full volume for 3D slices), spanning fine coronary trees and stenosis in X-ray angiography (left), cardiac chambers, masses, and scar defects in echocardiography and ultrasound (center), and volumetric gliomas and tumors in 3D MRI and CT (right). Shown are representative test cases.}
  \label{fig:qualitative}
\end{figure*}

\subsection{Parity holds mask by mask}
\label{sec:build_mask_by_mask}
On 2D tasks, parity with the specialist holds mask by mask as well (\Cref{tab:build_results}, right; \Cref{fig:qualitative}). \textbf{79\%} of all 2D test masks come within 0.05 Dice of nnU-Net or better (exceeding 50\% on every task), and the agent ties or outperforms nnU-Net on \textbf{47\%} of individual masks. Reliability likewise matches or exceeds the specialist: the agents' failure rate is lower than nnU-Net's (\textbf{24.0\%} vs.\ 25.8\%, tying or lower on six of nine tasks), while closely tracking nnU-Net on the hardest tail cases (the worst 10\% of masks) within 0.013 Dice (0.259 vs.\ 0.272).

\subsection{Nothing here is specific to segmentation}
\label{sec:build_nothing_specific}
Reaching specialist-level performance required nothing built specifically for segmentation. Even under Setting~1 (classical methods only, no neural networks), the agents' best programs already reach \textbf{82\%} of nnU-Net's 2D Dice. Adding PyTorch and MONAI from scratch (Setting~2) lifts 2D Dice by $+0.113$ on average, improving all 22 matched comparisons: the agents know how to engineer deep learning solutions when given the primitives, and they write that training code just as they write any other software. The capability stems directly from general-purpose coding. Handing the specialist's own tools back changes little: across six 2D tasks with Claude Opus~5.5, adding MedSAM and MedSAM2 with fine-tuning to the from-scratch toolbox (Setting~4) shifts Dice by only $+0.005$. Nor is this capability tied to a single model or to memorized benchmarks. All four agents across two vendors successfully built working segmenters. On 2D tasks, even the weakest single Setting~2 run across any agent reaches 0.558 Dice (\textbf{92\%} of nnU-Net), spanning a tight band of \textbf{92\%, 95\%, and 99\%} from weakest to median to best. Moreover, on 2D tasks from datasets released strictly after the agents' training cutoffs, the median run reaches \textbf{96\%} of nnU-Net (vs.\ 92\% on earlier datasets), ruling out benchmark memorization.

\subsection{A CPU, a few hours, a few dollars}
\label{sec:build_cost}

Treating segmentation as a coding task fundamentally alters the economics of building a segmenter. The median build takes 1.8\,h in Setting~1 and 5.8\,h in Setting~2 on a CPU, costing roughly \textbf{\$2.70} per Claude session at standard API rates. Training the GPU nnU-Net reference took a median of \textbf{18.9\,h} per task; every Setting~2 build session, training included, finishes within 7.3\,h---faster than the shortest nnU-Net training (11.1\,h). A new task thus shifts from a specialist team and GPU cluster to a task description and an afternoon on a CPU. Volumetric 3D segmentation is where this CPU budget binds: training 3D nnU-Net alone required a median of 27.0 GPU-hours per task, yet within an 8\,h CPU budget the best from-scratch 3D networks still reach \textbf{85\%} of its Dice and 4\,h classical pipelines reach \textbf{88\%} (\Cref{tab:build_results}), adapting naturally to the available compute.

Because the resulting segmenter is agent-written code, we next ask whether the same agent can improve its own outputs. In runs reaching a mean Dice of 0.7 or higher, the remaining headroom concentrates in just 8.1\% of masks---motivating the case-by-case review in the next section.
\section{Self-improvement: bounded alone, specialist-beating with a signal}
\label{sec:improve}
If a segmenter is a program written by an agent, it can be improved like code: revise a version, and check whether it got better. During development, this check comes from labeled validation data; on unseen test cases, there is no reference. Below, we show that visual inspection alone yields useful initial repairs before hitting a strict ceiling, whereas supplying a quality signal unlocks far larger gains that push the agents past GPU-trained nnU-Net.

\subsection{Looking alone improves every dimension}
\label{sec:self_review}

\begin{figure}[t]
  \centering
  \includegraphics[width=\linewidth]{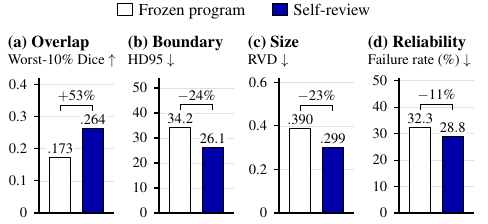}
\caption{\textbf{Visual self-review alone improves all four metric dimensions across all 91 runs.} Each panel compares the frozen program against reference-free Self-review: \textbf{(a)}~overlap on the hardest decile (worst-10\% mean Dice, \textbf{$+$53\%}), \textbf{(b)}~boundary error (median HD95, \textbf{$-$24\%}), \textbf{(c)}~size error (median RVD, \textbf{$-$23\%}), and \textbf{(d)}~failure rate (\textbf{$-$11\%} relative). Even without a reference, looking alone yields consistent gains, most of all on the hardest masks.}
\label{fig:self_review}
\end{figure}

In reference-free Self-review (\Cref{sec:design_self_review}), visual inspection alone improves every dimension of the metric panel across all 91 runs (\Cref{fig:self_review}), with the largest gains on the hardest cases: mean Dice on the worst 10\% of masks jumps from 0.173 to \textbf{0.264} (\textbf{+53\%}, improving in 73 of 91 runs). Boundaries tighten as median HD95 drops from 34.2 to \textbf{26.1} (\textbf{$-$24\%}, improved in 66 runs), size error falls as median RVD drops from 0.390 to \textbf{0.299} (\textbf{$-$23\%}, improved on 12 of 13 tasks), overall failure rate drops from 32.3\% to \textbf{28.8\%}, and mean Dice rises by 0.027.

Mask by mask across all 12,336 evaluated masks, twice as many masks improve (1,985) as degrade (984), and three failed masks are rescued for every mask pushed into failure (569 vs.\ 185; 344 vs.\ 111 crossing 0.8 Dice). On individual tasks, visual review alone cuts HD95 in half and slashes failures: on CARDIAG vessels, HD95 drops from 44.9 to 22.0 and failure rate from 10.3\% to 1.0\%; on CardioMassTTE-1151, HD95 drops from 46.0 to 20.4 and failure rate from 35.2\% to 18.0\%. Mechanistically, Self-review works by sharpening precision (rising from 0.610 to \textbf{0.654}, improving on eight tasks and decreasing on \textbf{none}): the agents reliably spot and prune gross over-segmentation in Round~1, but progress stalls immediately thereafter (\Cref{fig:round_curves}).

\subsection{With a quality signal, the agents go past the specialist}
\label{sec:true_dice}

\begin{figure}[t]
  \centering
  \includegraphics[width=\linewidth]{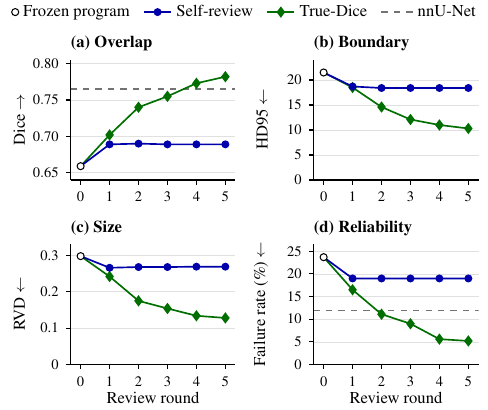}
\caption{\textbf{Self-review plateaus after Round~1, whereas a quality signal drives monotonic gains past nnU-Net.} Round-by-round trajectories averaged across the 14 matched pairs from Round~0 (frozen program) through Round~5 across \textbf{(a)}~overlap (Dice), \textbf{(b)}~boundary error (HD95), \textbf{(c)}~size error (RVD), and \textbf{(d)}~failure rate (dashed lines: GPU-trained nnU-Net). Guided by True-Dice, the agents improve every round---falling below nnU-Net's failure rate at Round~2, surpassing nnU-Net's Dice at Round~4, and reaching \textbf{0.790} final selected Dice (\Cref{tab:matched_pairs}).}
\label{fig:round_curves}
\end{figure}

\begin{table*}[t]
\caption{\textbf{Performance across the 14 matched pairs.} SR = Self-review, TD = True-Dice (\Cref{sec:design_matched}). \textbf{Bold} TD values surpass GPU-trained nnU-Net on the same masks. Rightmost columns list initially failed masks ($\text{Dice} < 0.5$) and the number rescued under SR and TD (bottom row: pair-averaged rescue rate). $^\dagger$3D task.}
  \label{tab:matched_pairs}
  \centering
  \footnotesize
  \setlength{\tabcolsep}{3.2pt}
  \begin{tabular}{@{}lccrccccccccrrr@{}}
    \toprule
    & & & & \multicolumn{4}{c}{Dice} & \multicolumn{2}{c}{HD95} & \multicolumn{2}{c}{RVD} & Failed & \multicolumn{2}{c}{Rescued} \\
    \cmidrule(lr){5-8} \cmidrule(lr){9-10} \cmidrule(lr){11-12} \cmidrule(l){14-15}
    Task & Setting & Agent & Masks & Program & SR & TD & nnU-Net & SR & TD & SR & TD & masks & SR & TD \\
    \midrule
    FeEcho4D & 1 & Opus & 40 & .627 & .651 & \textbf{.727} & .694 & 3.27 & 2.12 & .201 & .139 & 11 & 4 & 10 \\
    FeEcho4D & 1 & Fable & 40 & .603 & .606 & \textbf{.720} & .694 & 3.57 & 2.02 & .430 & .118 & 15 & 0 & 13 \\
    CardioMassTTE-1151 & 1 & Opus & 20 & .634 & .752 & \textbf{.838} & .813 & 17.1 & 16.2 & .139 & .076 & 4 & 3 & 4 \\
    CardioMassTTE-1151 & 1 & Fable & 20 & .624 & .737 & \textbf{.852} & .813 & 26.2 & 11.0 & .294 & .055 & 5 & 4 & 5 \\
    FeEcho4D & 2 & Opus & 40 & .751 & .751 & \textbf{.786} & .694 & 1.47 & 1.25 & .105 & .060 & 2 & 0 & 2 \\
    CardioMassTTE-1151 & 2 & Opus & 20 & .795 & .830 & \textbf{.866} & .813 & 14.7 & 12.2 & .152 & .055 & 2 & 1 & 2 \\
    CARDIAG vessels & 1 & Opus & 20 & .809 & .825 & .859 & .895 & 20.9 & 15.8 & .109 & .049 & 0 & -- & -- \\
    ARCADE stenosis & 2 & Opus & 20 & .414 & .444 & \textbf{.687} & .630 & 81.4 & 25.6 & .567 & .293 & 12 & 1 & 11 \\
    CT-PEGCT-Diag$^\dagger$ & 2 & Opus & 20 & .644 & .649 & \textbf{.826} & .784 & 13.2 & 5.19 & .236 & .081 & 7 & 1 & 7 \\
    ARCADE segments & 1 & Opus & 143 & .344 & .358 & \textbf{.582} & .497 & 59.0 & 31.0 & .587 & .288 & 88 & 8 & 52 \\
    WAW-GliomAI post-op$^\dagger$ & 1 & Opus & 99 & .619 & .631 & .708 & .736 & 6.40 & 5.05 & .256 & .147 & 25 & 1 & 10 \\
    CesareanScarDefect & 1 & Opus & 20 & .644 & .702 & \textbf{.814} & .829 & 8.55 & 6.66 & .475 & .145 & 3 & 2 & 3 \\
    WAW-GliomAI pre-op$^\dagger$ & 1 & Opus & 82 & .833 & .843 & .889 & .892 & 2.00 & 1.41 & .104 & .038 & 8 & 1 & 4 \\
    CAMUS & 1 & Opus & 60 & .886 & .875 & .907 & .923 & 4.88 & 3.70 & .103 & .045 & 0 & -- & -- \\
    \midrule
    \textbf{Mean} &  &  &  & \textbf{.659} & \textbf{.689} & \textbf{.790} & \textbf{.765} & \textbf{18.7} & \textbf{9.9} & \textbf{.269} & \textbf{.113} &  & 30\% & \textbf{85\%} \\
    \bottomrule
  \end{tabular}
\end{table*}

To measure how much further the same agents can go when guided by reliable feedback, we evaluate the True-Dice upper bound across all 14 matched pairs (\Cref{sec:design_matched}; \Cref{tab:matched_pairs}, \Cref{fig:round_curves}). With a quality signal, every single pair improves substantially: mean Dice rises from 0.659 (frozen) and 0.689 (Self-review) to \textbf{0.790} under True-Dice---a net gain of \textbf{$+0.101$} over Self-review, with the 95\% confidence interval strictly above zero for every pair. Across 280 paired case comparisons, True-Dice outperforms Self-review 246 times vs.\ 14 ($p = 6\times 10^{-56}$, sign test), and the pair-averaged rescue rate for failed masks leaps from 30\% to \textbf{85\%}.

Crucially, True-Dice pushes the agents' mean Dice (\textbf{0.790}) past GPU-trained nnU-Net on the exact same masks (0.765), propelling \textbf{eight} of the 13 pairs that started below nnU-Net past the specialist model. Mask by mask, the share of masks within 0.05 Dice of nnU-Net (or better) climbs from 46\% (frozen) and 49\% (Self-review) to \textbf{75\%}, while the share tying or beating nnU-Net nearly doubles from 20\% and 23\% to \textbf{43\%}. Remarkably, all initial programs were built on a CPU, and \textbf{10 of the 14 pairs} started from Setting~1 classical pipelines---meaning a CPU-built classical program paired with a quality signal outperforms a GPU-trained deep neural network.

Although told only scalar Dice, every geometric and reliability metric improves in lockstep: across all 14 pairs, mean HD95 drops by \textbf{47\%} vs.\ Self-review (18.7 to \textbf{9.9}) and RVD by \textbf{58\%} (0.269 to \textbf{0.113}). Failure rate falls by \textbf{78\%} (19.0\% to \textbf{4.1\%})---one-third that of nnU-Net (11.9\%)---while worst-10\% Dice triples from 0.208 (frozen) and 0.398 (Self-review) to \textbf{0.641}. Both recall (0.728 $\rightarrow$ 0.811) and precision (0.717 $\rightarrow$ 0.788) rise together, confirming genuinely superior boundaries and volumes across both 2D (HD95: 21.9 $\rightarrow$ 11.6) and 3D tasks (HD95: 7.2 $\rightarrow$ 3.9). These gains accumulate monotonically across all five rounds (\Cref{fig:round_curves}; \Cref{fig:overview}c). Decomposing the $+0.101$ Dice gain over Self-review shows that \textbf{$+0.089$} comes from \textbf{generating better masks} and only $+0.012$ from selecting among existing attempts; even the best candidate across all Self-review rounds averages only 0.702 Dice.

\subsection{A signal is worth more than a model}
\label{sec:signal_vs_model}

Starting from the same frozen programs on the same masks (0.659 Dice), replacing the agent's program with a GPU-trained nnU-Net yields \textbf{$+0.106$ Dice}; keeping the agent's CPU program and supplying a quality signal yields \textbf{$+0.131$ Dice}. The capacity to repair and surpass specialist segmentations is already inside general-purpose coding agents; what unlocks it is a quality signal.
\section{What supplies the quality signal?}
\label{sec:bottleneck}
True-Dice shows that a reliable quality signal enables coding agents to surpass specialist models, yet computing true Dice requires a reference mask unavailable at deployment. What, then, can supply this signal on unseen cases? We test three reference-free candidates in turn: the agent's own visual self-ratings, additional rounds of revision effort, and a quality estimator trained by the agent itself.

\subsection{Candidate 1: visual self-ratings}
\label{sec:eyes}
Section~5.1 showed that visual inspection repairs gross errors in Round~1 before stalling (\Cref{fig:round_curves}). To diagnose why vision cannot guide further rounds, we audit the 0--100 visual self-ratings the agents recorded at every round of Self-review. Across the 2,107 cases where the agent modified its prediction, it assigned a higher self-rating to its revision in 1,908 cases (\textbf{91\%}); true Dice actually increased in 1,048 cases (\textbf{50\%})---no better than a coin flip (\Cref{fig:overview}d). The visual self-rating does carry global information across cases: its rank correlation with true Dice is \textbf{0.47}, showing that the agent can tell which cases are hard. Iterative improvement, however, requires judging whether a specific revision on the \emph{same} case made the mask better or worse. On this within-case judgment, directional accuracy is \textbf{66.7\%} (where random guessing is 50\%), holding across both vendors (69.6\% for Claude, 55.4\% for GPT), and selecting the highest self-rated candidate yields the exact same performance as the agent's final choice. Thus, the agent's eyes can rank cases by difficulty, but cannot reliably rank its own revisions by quality: \textbf{seeing is not verifying}.

\subsection{Candidate 2: additional revision rounds}
\label{sec:more_rounds}
The second candidate is sheer effort. Because agents under True-Dice revise across more rounds, Forced-rounds (\Cref{sec:design_matched}) tests whether extra effort suffices by requiring all five rounds without external scores across four tasks. Over Rounds~2--5, revision without a signal changes mean Dice by $+0.004$ (vs.\ $+0.000$ under Self-review); with True-Dice, the same four rounds yield \textbf{$+0.119$ Dice}, consistent across all four tasks. Every other dimension tells the same story: Rounds~2--5 under Forced-rounds vs.\ True-Dice reduce HD95 by 2.7 vs.\ \textbf{17.9}, reduce RVD by 0.013 vs.\ \textbf{0.149}, shift failure rate by $+1.9$ vs.\ \textbf{$-$20.0} percentage points, and lift worst-10\% Dice by $+0.016$ vs.\ \textbf{$+0.202$}. The sharpest contrast appears on failed masks: four extra rounds without a signal increase active failures from 22 at Round~1 to 25 and rescue only 10 of 34 initially failed masks (vs.\ 9 under Self-review), whereas True-Dice cuts active failures from 20 at Round~1 to \textbf{2} and rescues \textbf{32} of 34. What drives improvement is the signal, not the number of rounds.

\subsection{Candidate 3: an agent-trained quality estimator}
\label{sec:self_critic}
The third is the most direct engineering approach: having the agent build the quality signal itself. Under Self-critic (\Cref{sec:design_matched}), the agent trains its own reference-free Dice estimator on labeled training/validation predictions and synthetic mask perturbations, then queries it at each review round. Across all five metrics---Dice, HD95, RVD, failure rate, and worst-10\% Dice---the median fraction of the Self-review-to-True-Dice gap closed by Self-critic is \textbf{0\%} (mean Dice 0.709 vs.\ 0.700 for Self-review and \textbf{0.798} for True-Dice; failure rate 16.8\% vs.\ 18.2\% and \textbf{1.1\%}). Out of 38 initially failed masks, Self-critic rescues 11, vs.\ 10 under Self-review and \textbf{36} under True-Dice. On CT-PEGCT-Diag, where Self-critic reaches 0.765 Dice, Forced-rounds without any estimator reaches 0.770 on the exact same cases, confirming that the gain does not come from the learned estimator.

The agents' training logs reveal why. When ranking all validation masks globally, the self-trained estimators achieve a high rank correlation with true Dice (\textbf{0.66 to 0.88}); yet when ranking competing revisions of the \emph{same} case, rank correlation drops to \textbf{0.25}. The estimators easily distinguish artificially corrupted masks from clean ones, but cannot distinguish subtle improvements from degradations among the frozen program's actual failure modes. This pinpoints a concrete requirement for future work: a useful quality estimator must discriminate between competing revisions of the same case, which high global correlation across cases does not guarantee.

Taken together, none of the three reference-free candidates supplies the needed within-case verification signal.
\section{Discussion}
\label{sec:discussion}

None of the three reference-free candidates in \Cref{sec:bottleneck} supplies the needed verification signal. We term this bottleneck the \emph{quality blind spot}:

\begin{keybox}
\textbf{Quality blind spot:} \emph{Without a reference, a coding agent can see which cases are hard, but cannot tell whether its own revision made a mask better or worse.}
\end{keybox}

\noindent\textbf{Segmentation inherits both halves of a coding task.}
Every behavior we observe here mirrors a known property of coding agents on software: just as software agents build working programs from specifications~\cite{jimenez2024swebenchlanguagemodelsresolve, yang2024sweagentagentcomputerinterfacesenable} and iterate to pass unit tests when executable feedback is given~\cite{yang2024sweagentagentcomputerinterfacesenable}, our agents build segmenters from scratch (reaching \textbf{99\%} of nnU-Net on 2D) and improve in every round under True-Dice (rescuing \textbf{85\%} of failed masks). Conversely, without external tests, software agents report ``fixed'' without knowing~\cite{huang2024largelanguagemodelsselfcorrect, kamoi-etal-2024-llms}, gain nothing from extra retries~\cite{huang2024largelanguagemodelsselfcorrect}, and cannot rely on self-written tests~\cite{kamoi-etal-2024-llms}---matching our findings on visual self-ratings (\textbf{91\%} believed vs.\ \textbf{50\%} actual), Forced-rounds ($+$0.004 vs.\ \textbf{{\boldmath$+$}0.119} Dice), and Self-critic (\textbf{0\%} median gap closed). Segmentation thus inherits both halves of a coding task, making the review experiments in the second half of this study just as much evidence for our title as the build experiments in the first half. Crucially, segmentation adds one finding that pure software could not test: even when the output is an image rendered directly before the agent's eyes, vision does not substitute for verification---\textbf{seeing is not verifying}.

\smallskip
\noindent\textbf{Where the next advance will come from.}
This quality blind spot is not specific to medical imaging: it is the general verification problem of how a coding agent confirms its own output. For segmentation, the next advance will therefore come from \emph{reference-free quality estimation}---a callable signal that a coding agent can query on demand, which is already worth more on the same masks than swapping in a GPU-trained specialist model (\Cref{sec:signal_vs_model}). Our matched design also provides a direct ruler for future work: any reference-free signal can be placed in the position of True-Dice to measure what fraction of the Self-review-to-True-Dice gap (from 0\% to 100\%) it closes. Crucially, \Cref{sec:self_critic} pinpoints what this ruler must test: discriminating better from worse among competing revisions of the \emph{same} case, rather than global correlation across cases.

\smallskip
\noindent\textbf{From a single repair to a self-improving loop.}
This study measures generation zero of a loop: building a program once and repairing its outputs once. Our results show what sets the full loop in motion. With a reliable quality signal, repair yields superior masks that can flow back into development as new training data and experience, letting the agent build the next version of the program. Self-improvement requires generation, evaluation, and revision; generation and revision are already in place, and evaluation is the switch that turns the loop on, governed by a simple rule: no stage grades itself, and a small held-out reference at the outer boundary serves as the anchor. Medical image segmentation is an ideal testbed for studying such agent self-improvement: its metrics are objective, reference annotations are compact, tasks are abundant, and each round is short, allowing the progress an agent claims and the progress it actually makes to be measured separately at every round.
\section{Conclusion}
\label{sec:conclusion}
Medical image segmentation is now just another coding task: given only a task description, labeled data, and a CPU, general-purpose coding agents build segmenters that reach 99\% of the Dice of a GPU-trained specialist on 2D tasks. The same agents can already repair their own predictions; with a reliable quality signal, they improve across overlap, boundary, size, and reliability, and surpass the specialist. The next advance will therefore come not from another segmentation model, but from overcoming this \emph{quality blind spot} via reference-free quality estimation that coding agents can call.
{
    \small
    \providecommand{\bibogonek}[1]{{\fontencoding{T1}\selectfont\k{#1}}}

}

\end{document}